\documentclass[sigconf]{acmart}
\usepackage{caption}
\usepackage{subcaption}
\usepackage{todonotes}

\setcopyright{none}
\copyrightyear{2022}
\acmYear{2022}
\acmDOI{}

\acmConference[ERN'22]{Accepted at the 2022 Emerging Researchers National (ERN) Conference in STEM}{due to COVID-19, the conference is postponed until 2023}{Washington, D.C.}
\acmBooktitle{Accepted at the 2022 Emerging Researchers National (ERN) Conference in STEM, Washington, D.C., due to COVID-19, the conference is postponed until 2023}
\acmPrice{}
\acmISBN{}

\makeatletter
\def\@copyrightspace{\relax}
\makeatother

\begin{document}

\title{Time Series Analysis of Blockchain-Based Cryptocurrency Price Changes}

\author{Jacques Fleischer}
\email{jacques.fleischer002@mymdc.net}
\affiliation{%
  \institution{Miami Dade College}
  \streetaddress{11011 SW 104th St}
  \city{Miami}
  \state{FL}
  \country{USA}
}

\author{Gregor von Laszewski}
\email{laszewski@gmail.com}
\orcid{0000-0001-9558-179X}
\affiliation{%
  \institution{University of Virginia}
  \streetaddress{Biocomplexity Institute\\
                Town Center Four\\
                994 Research Park Boulevard}
  \city{Charlottesville}
  \state{VA}
  \country{USA}
}

\author{Carlos Theran}
\email{carlos.theran@famu.edu}
\affiliation{%
  \institution{Florida A\&M University}
  \streetaddress{1601 S Martin Luther King Jr Blvd}
  \city{Tallahassee}
  \state{FL}
  \country{USA}
}

\author{Yohn Jairo Parra Bautista}
\email{yohn.parrabautista@famu.edu}
\affiliation{%
  \institution{Florida A\&M University}
  \streetaddress{1601 S Martin Luther King Jr Blvd}
  \city{Tallahassee}
  \state{FL}
  \country{USA}
}

\renewcommand{\shortauthors}{J.P. Fleischer, G. von Laszewski, C. Theran, Y. J. Parra Bautista.}

\begin{abstract}
  In this paper we apply neural networks and Artificial Intelligence (AI) to historical records of high-risk cryptocurrency coins to train a prediction model that guesses their price. This paper's code contains Jupyter notebooks, one of which outputs a timeseries graph of any cryptocurrency price once a CSV file of the historical data is inputted into the program. Another Jupyter notebook trains an LSTM, or a long short-term memory model, to predict a cryptocurrency's closing price. The LSTM is fed the close price, which is the price that the currency has at the end of the day, so it can learn from those values. The notebook creates two sets: a training set and a test set to assess the accuracy of the results.

The data is then normalized using manual min-max scaling so that the model does not experience any bias; this also enhances the performance of the model. Then, the model is trained using three layers— an LSTM, dropout, and dense layer—minimizing the loss through 50 epochs of training; from this training, a recurrent neural network (RNN) is produced and fitted to the training set. Additionally, a graph of the loss over each epoch is produced, with the loss minimizing over time. Finally, the notebook plots a line graph of the actual currency price in red and the predicted price in blue. The process is then repeated for several more cryptocurrencies to compare prediction models. The parameters for the LSTM, such as number of epochs and batch size, are tweaked to try and minimize the root mean square error.
\end{abstract}



\maketitle

\section{Introduction}

Blockchain is {\em an open, distributed ledger} which records
transactions of cryptocurrency. Systems in blockchain are
decentralized, which means that these transactions are shared and
distributed among all participants on the blockchain for maximum
accountability. Furthermore, this new blockchain technology is
becoming an increasingly popular alternative to mainstream
transactions through traditional banks \cite{c2}. These transactions
utilize blockchain-based {\em cryptocurrency}, which is a popular
investment of today's age, particularly in Bitcoin. However, the
U.S. Securities and Exchange Commission warns that high-risk
accompanies these investments \cite{c1}.

Artificial Intelligence (AI) can be used to predict the prices'
behavior to avoid cryptocurrency coins' severe volatility that can
scare away possible investors \cite{c3}. AI and blockchain technology
make an ideal partnership in data science; the insights generated from
the former and the secure environment ensured by the latter create a
goldmine for valuable information. For example, an up-and-coming
innovation is the automatic trading of {\em digital investment assets}
by AI, which will hugely outperform trading conducted by humans
\cite{c5}. This innovation would not be possible without the
construction of a program which can pinpoint the most ideal time to
buy and sell. Similarly, AI is applied in this experiment to predict
the future price of cryptocurrencies on a number of different
blockchains, including the Electro-Optical System and Ethereum.

Long short-term memory (LSTM) is a neural network (form of AI) which
ingests information and processes data using a {\em gradient-based
  learning algorithm} \cite{c6}. This creates an algorithm that
improves with additional parameters; the algorithm {\em learns} as it
ingests. LSTM neural networks will be employed to analyze pre-existing
price data so that the model can attempt to generate the future price
in varying timetables, such as ten days, several months, or a year
from the last date. This innovation could provide as a boon for
insights into investments with potentially great returns; it could
also contribute to a positive cycle of attracting investors to a coin,
which results in a price increase, which repeats. The main objective
is to provide insights for investors on an up-and-coming product:
cryptocurrency.

\section{Datasets}

\begin{figure}[htb]
\includegraphics[width=\columnwidth]{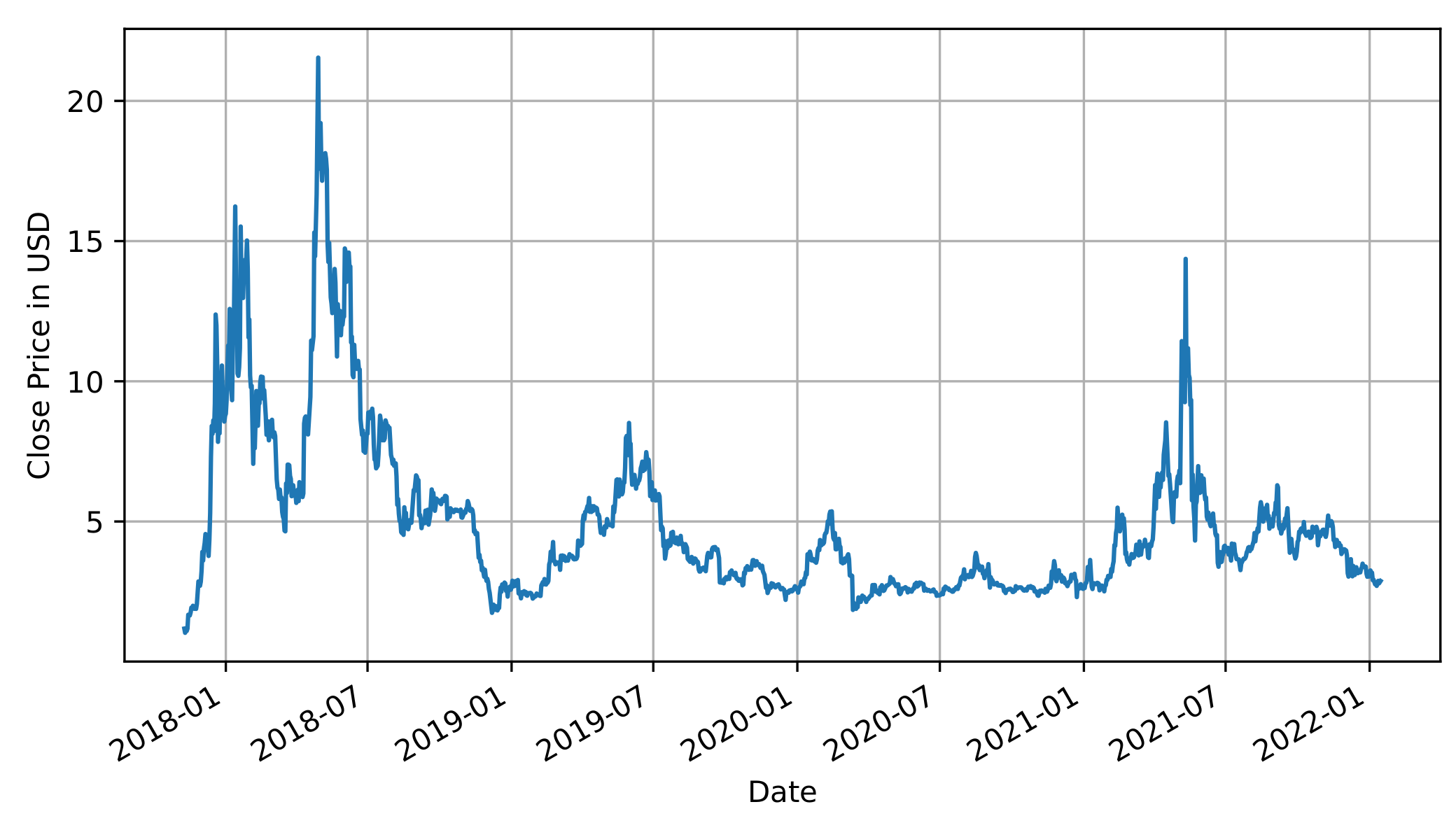}

\caption{Line graph of EOS price from 9 November 2017 to 13 January
  2022. Generated using yfinance-lstm.ipynb \cite{c13} located in
  project/code, utilizing price data from Yahoo Finance \cite{c4}.}

\label{fig:eos-price}
\end{figure}

This paper utilizes yfinance, a Python module which downloads the
historical prices of a cryptocurrency from the first day of its
inception to whichever day the program is executed. For example, the
Yahoo Finance page for EOS-USD is the source for Figure
\ref{fig:eos-price} \cite{c4}. Figure \ref{fig:eos-price} shows the
historical data on a line graph when the program receives EOS-USD as
an input.

\section{Architecture}

\begin{figure}[htb]
\includegraphics[width=\columnwidth]{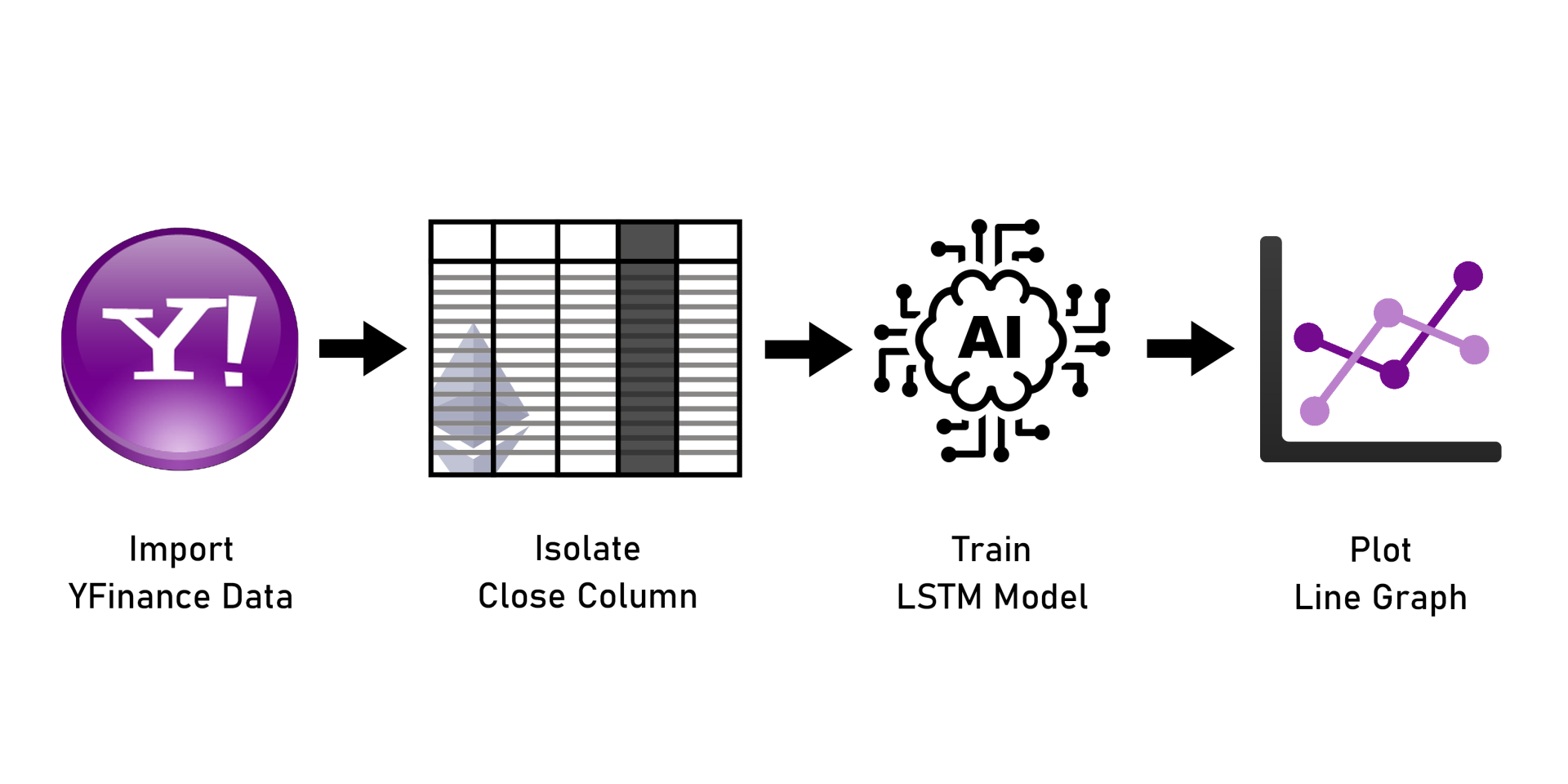}
\caption{The process of producing LSTM timeseries based on cryptocurrency price.}
\label{fig:arch-process}
\end{figure}

This program undergoes the four main phases outlined in Figure
\ref{fig:arch-process}, which are: retrieving data from Yahoo Finance
\cite{c4}, isolating the Close prices (the price the cryptocurrency
has at the end of each day), training the LSTM to predict Close
prices, and plotting the prediction model, respectively.

\section{Implementation}

Initially, this program was meant to scrape prices using the
BeautifulSoup Python module; however, slight changes in a financial
page's website caused the code to break. Alternatively, Kaggle offered
historical datasets of cryptocurrency, but they were not up to
date. Thus, the final method of retrieving data is from Yahoo Finance
through the yfinance Python module, which returns the coins' price
from the day to its inception to the present day.

The code is inspired from Towards Data Science articles by Serafeim
Loukas \cite{c7} and Viraf \cite{c11}, who explore using LSTM to
predict stock timeseries. This program contains adjustments and
changes to their code so that cryptocurrency is analyzed instead. We
opt to use LSTM (long short-term memory) to predict the price because
it has a memory capacity, which is ideal for a timeseries data set
analysis such as cryptocurrency price over time. LSTM can remember
historical patterns and use them to inform further predictions; it can
also selectively choose which datapoints to use and which to disregard
for the model \cite{c8}. For example, this experiment's code isolates
only the close values to predict them and nothing else.

Firstly, the code asks the user for the ticker of the cryptocurrency
that is to be predicted, such as EOS-USD or BTCUSD. A complete list of
acceptable inputs is under the Symbol column at the Yahoo Finance list
of cryptocurrencies \cite{c17} but theoretically, the program should
be able to analyze traditional stocks as well.

Then, the program downloads the historical data for the corresponding
coin through the yfinance Python module \cite{c18}. The data must go
through normalization for simplicity and optimization of the
model. Next, the Close data (the price that the currency has at the
end of the day, everyday since the coin's inception) is split into two
sets: a training set and a test set, which are further split into
their own respective x and y sets to guide the model through training.

\begin{figure}[htb]
\includegraphics[width=0.5\columnwidth]{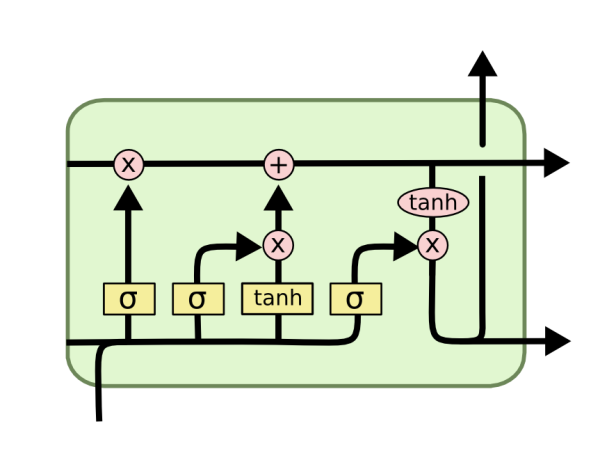}
\caption{Visual depiction of one layer of long short-term memory \cite{c9}.}
\label{fig:alstm}
\end{figure}

The training model is run through a layer of long short-term memory,
as well as a dropout layer to prevent overfitting and a dense layer to
give the model a memory capacity. Figure \ref{fig:alstm} showcases the
setup of the LSTM layer.

The entire program which performs all of the aforementioned steps can
be found on GitHub \cite{c13}.

\begin{figure}[htb]
\includegraphics[width=\columnwidth]{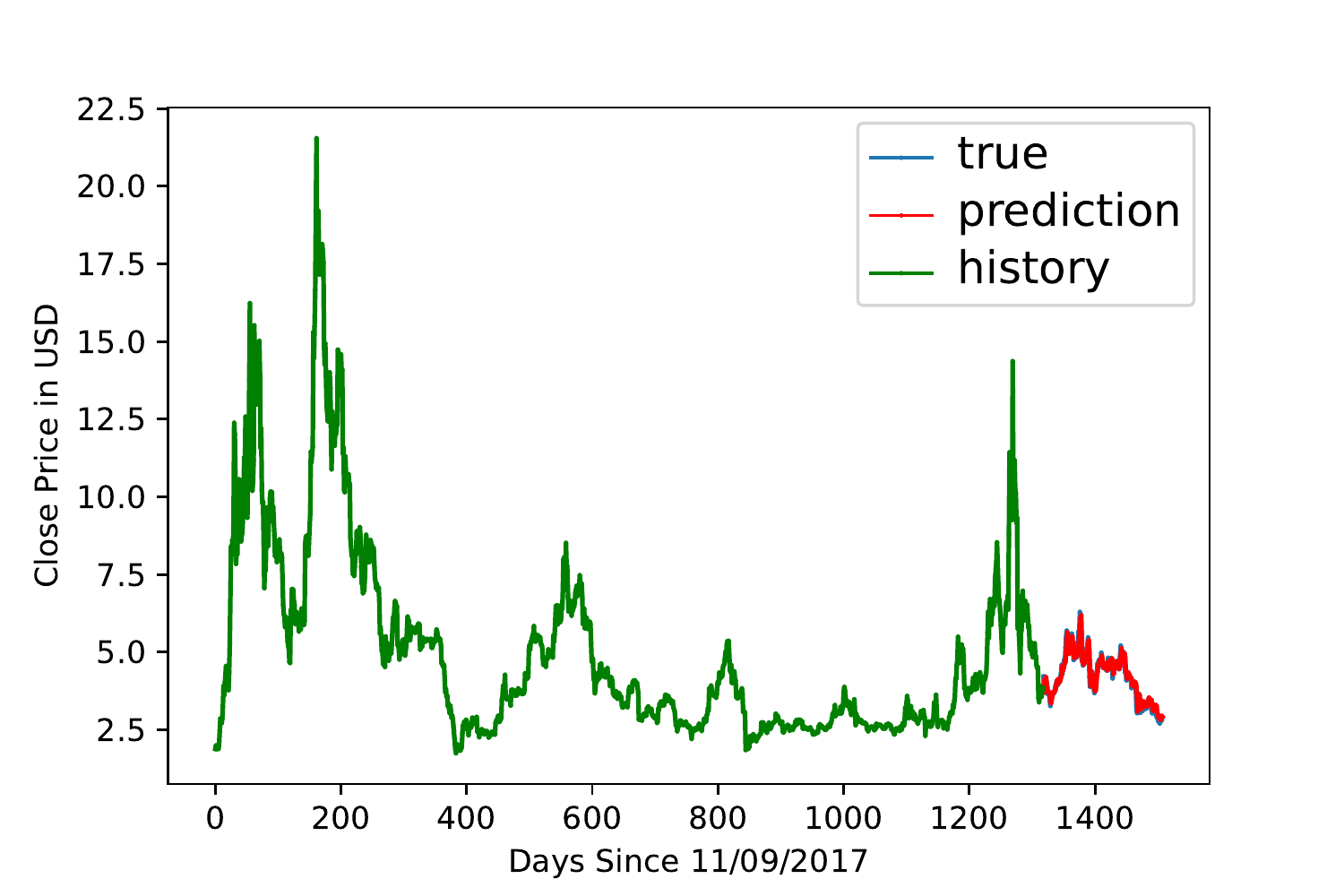}
\caption{EOS-USD price overlayed with the latest 200 days predicted by LSTM.}
\label{fig:prediction-model}
\end{figure}

\begin{figure}[htb]
\includegraphics[width=\columnwidth]{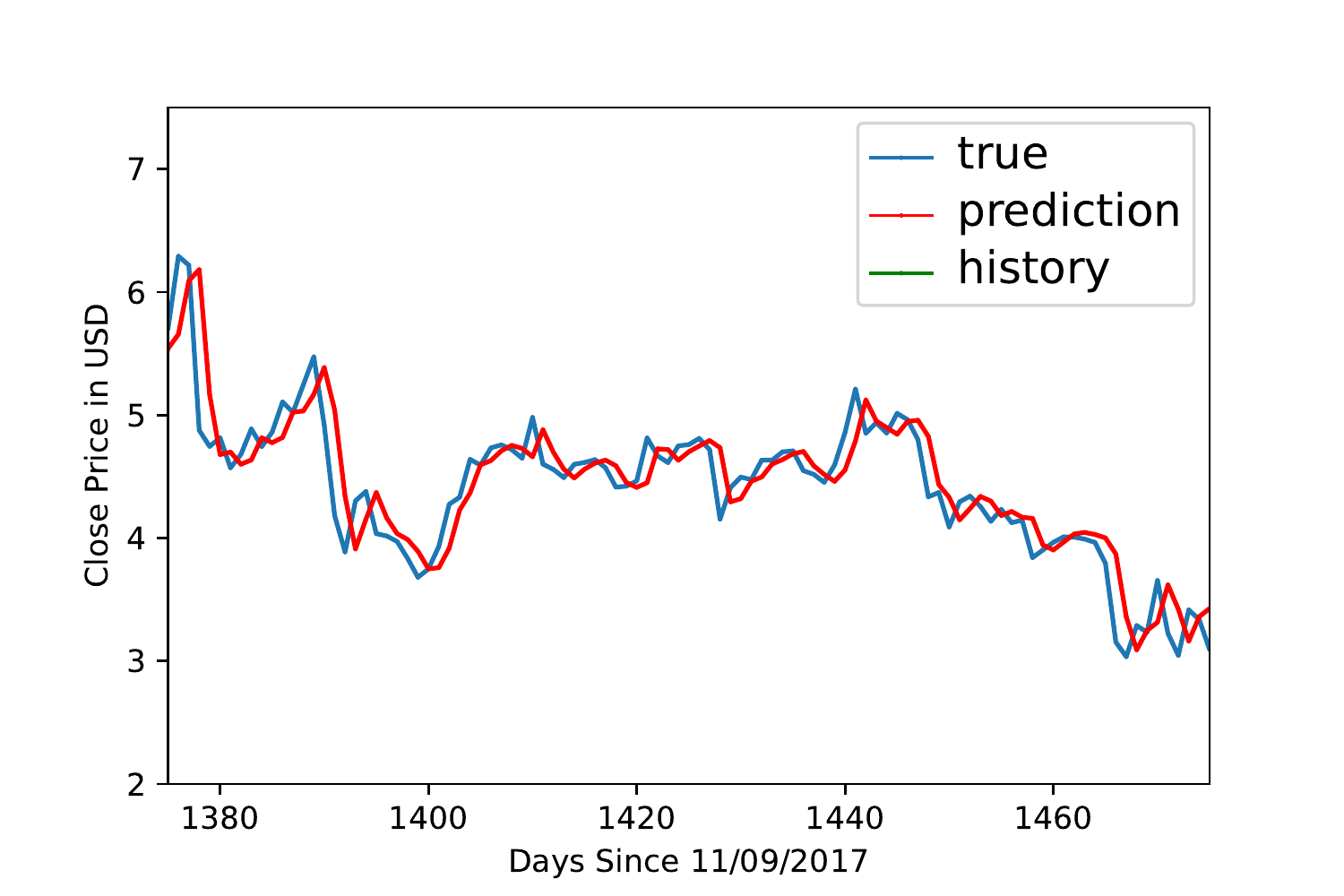}

\caption{Zoomed-in graph (same as Figure \ref{fig:prediction-model} but scaled x and y-axis for readability.}
\label{fig:zoomed}
\end{figure}

\begin{figure}[htb]
\includegraphics[width=\columnwidth]{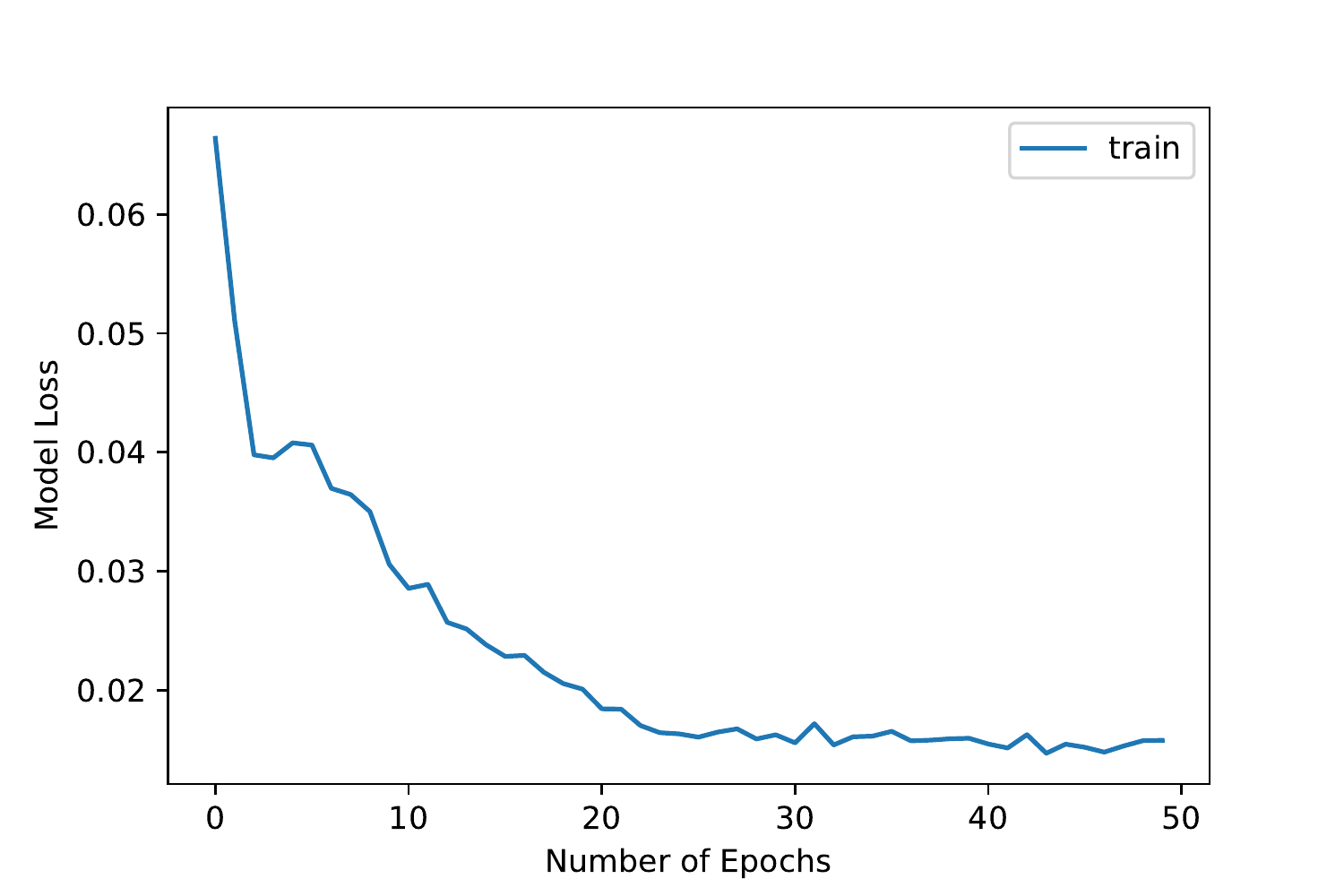}

\caption{Line graph of model loss over the number of epochs the
  prediction model completed using EOS-USD data set.}

\label{fig:loss}
\end{figure}

\begin{figure*}[htb]
\centering
     \begin{subfigure}[b]{0.49\textwidth}
         \centering
         \includegraphics[width=\textwidth]{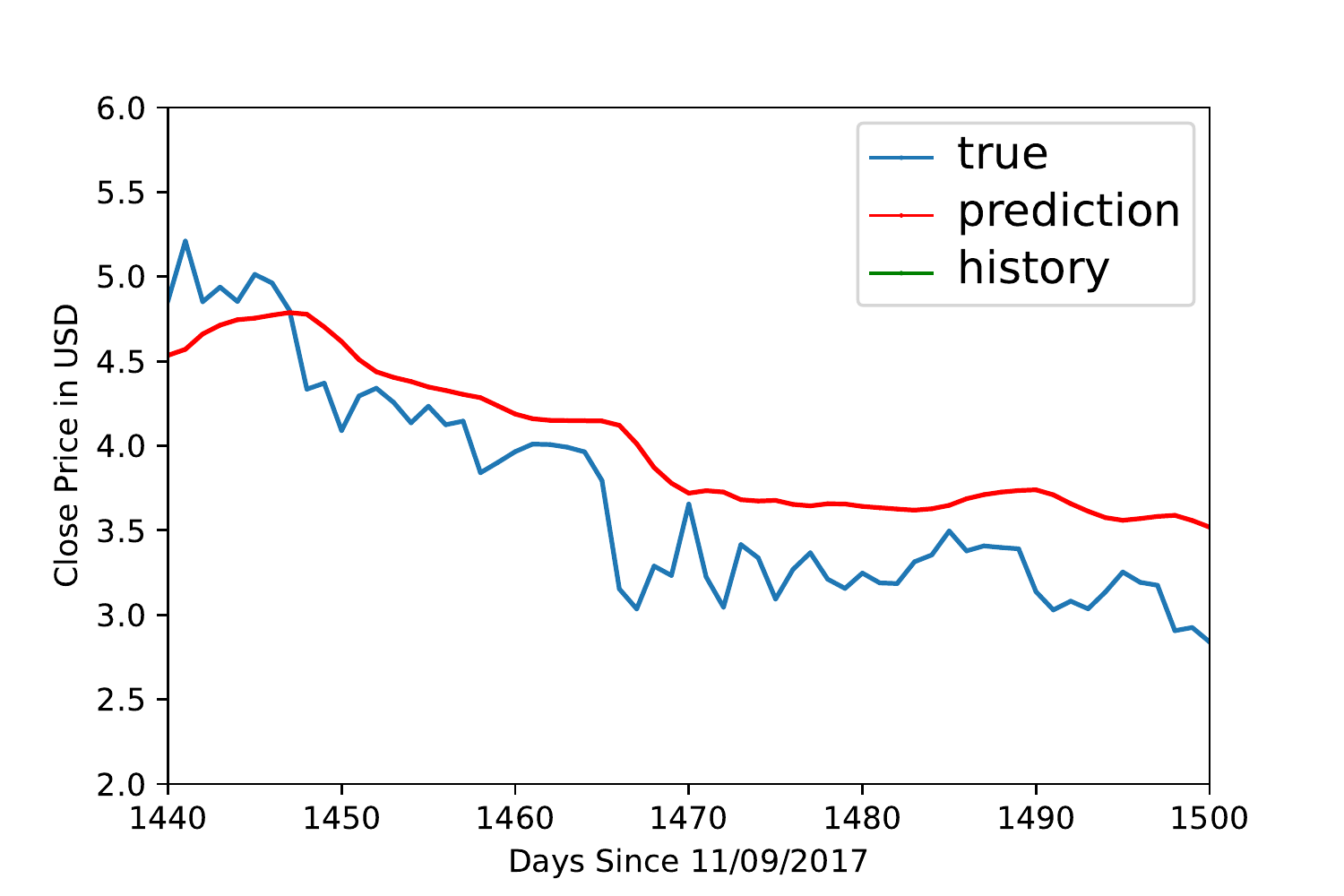}
         \caption{5 Epochs}
         \label{fig:epochs-5}
     \end{subfigure}
     \hfill
    \begin{subfigure}[b]{0.49\textwidth}
         \centering
         \includegraphics[width=\textwidth]{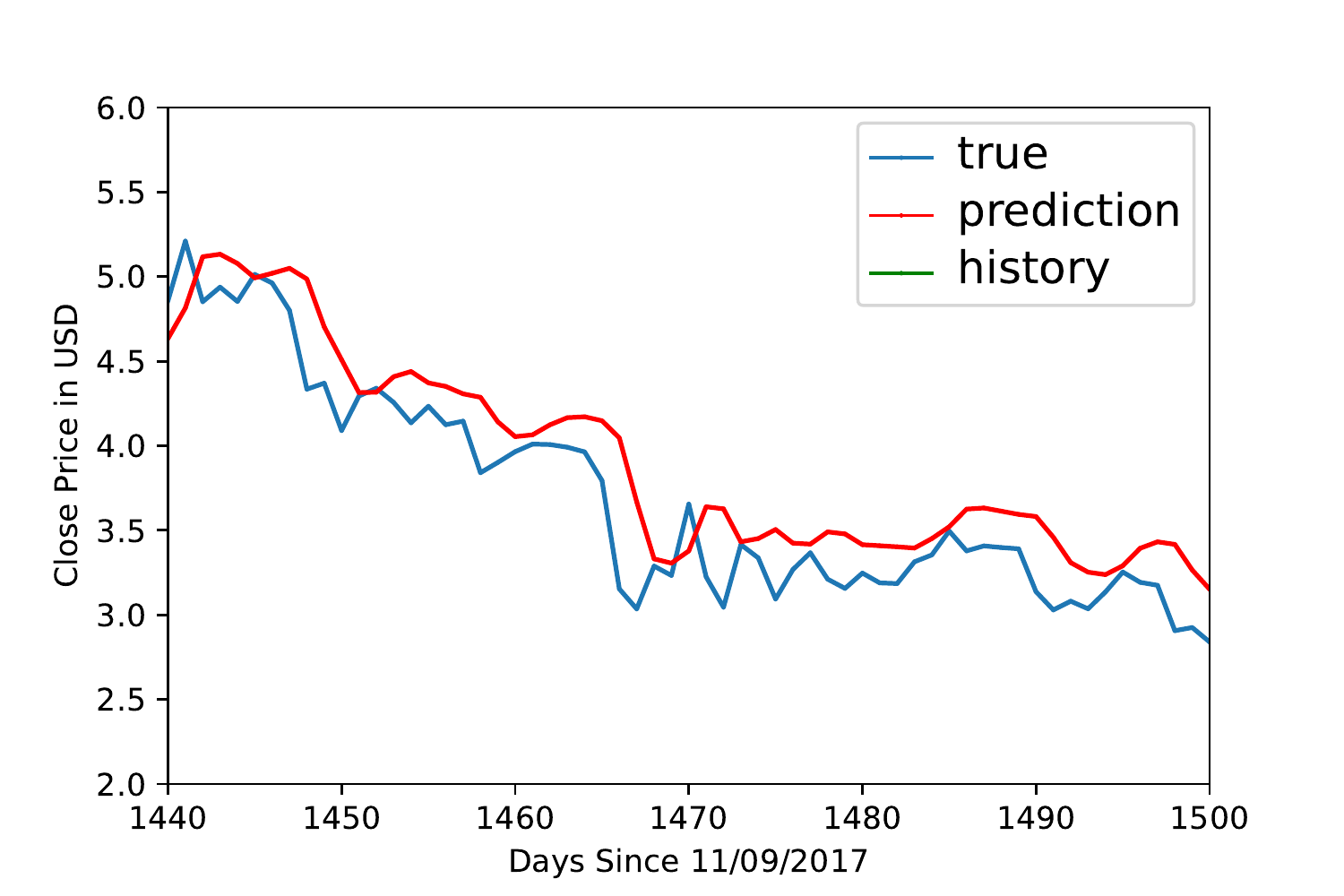}
         \caption{25 Epochs}
         \label{fig:epochs-25}
     \end{subfigure}
     \\
     \begin{subfigure}[b]{0.49\textwidth}
         \centering
         \includegraphics[width=\textwidth]{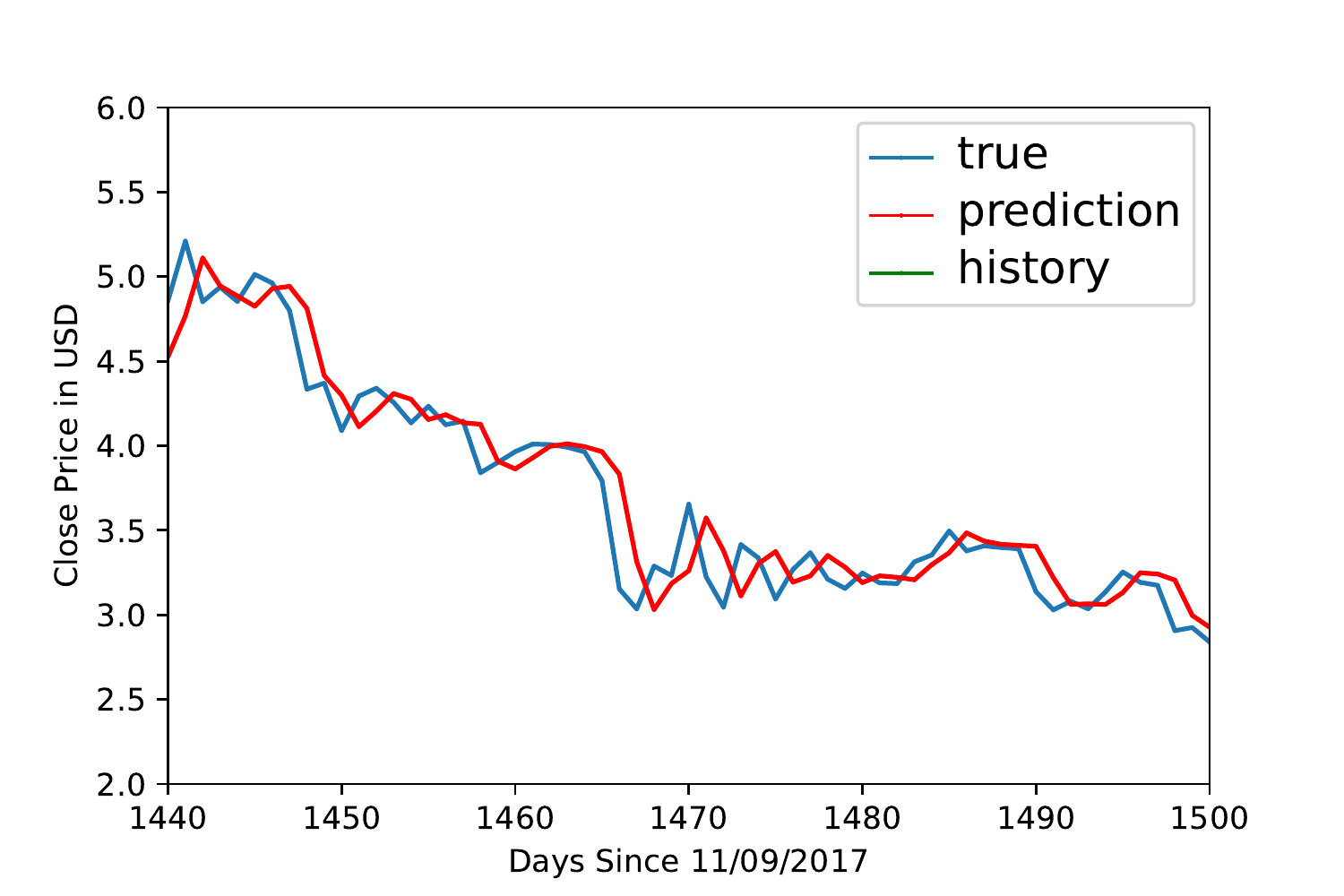}
         \caption{50 Epochs}
         \label{fig:epochs-50}
     \end{subfigure}
     \hfill
    \begin{subfigure}[b]{0.49\textwidth}
         \centering
         \includegraphics[width=\textwidth]{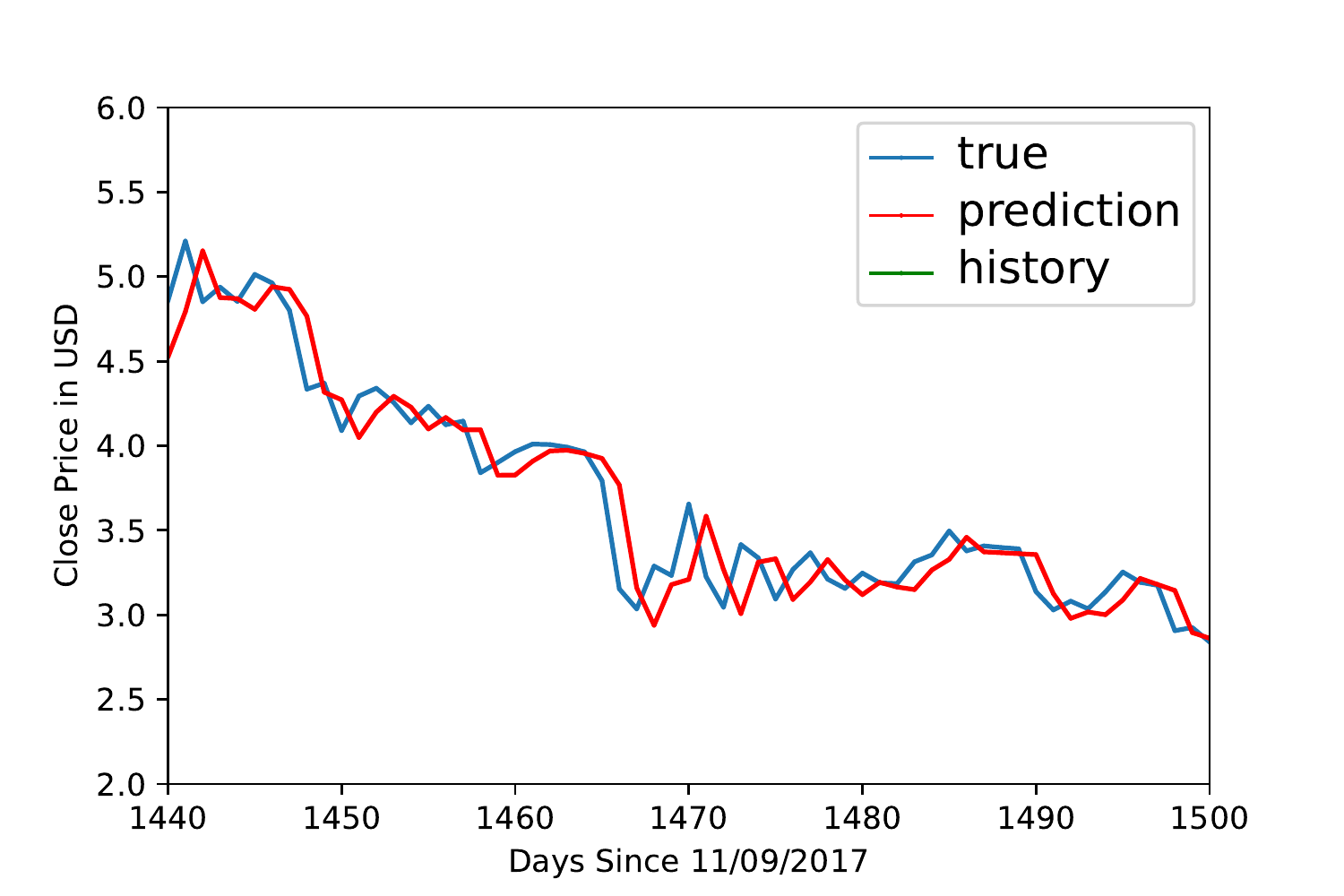}
         \caption{100 Epochs}
         \label{fig:epochs-100}
     \end{subfigure}

\caption{Effect of EOS-USD prediction model based on the number of epochs completed.}
\label{fig:epoch}
\end{figure*}

Figures \ref{fig:prediction-model} through \ref{fig:epoch} use the
EOS-USD data set from November 9th, 2017 to January 13th, 2022;
furthermore, these figures were all produced within the yfinance-lstm
Jupyter Notebook \cite{c13}. Within the trained model, only the last
200 days are predicted so that the model can analyze the preexisting
data prior to the 200 days for the sake of training.

After training through 50 epochs, the program generated Figure
\ref{fig:prediction-model}, a line graph of the prediction model.

During training, the number of epochs can affect the model
loss. According to Figures \ref{fig:loss} and \ref{fig:epoch}, the
loss starts to minimize around the 25th epoch of training. The greater
the number of epochs, the sharper and more accurate the prediction
becomes, but it does not vastly improve after around the 25th epoch.

The number of training epochs can affect the Root Mean Squared Error
of the model, which details how close the prediction line is to the
real, historical Close prices in United States Dollars (USD). As
demonstrated in Table \ref{tab:epochvsrmse}, more epochs lessens the
Root Mean Squared Error (but the change becomes negligible after 25
epochs).

Figure \ref{fig:epoch} also shows the impact that epochs have on
accuracy. Figure \ref{fig:epoch} contains two lines: a blue line for
the actual price of the EOS coin, and a red line for the model's
prediction of the price. As the number of epochs increases, the
prediction becomes more and more accurate to the actual price that the
cryptocoin was valued at on the market. In Figure \ref{fig:epoch}, the
green "history" line is not shown because the graph is zoomed in to
the later prediction phase, where the historical price data becomes
the blue line instead of green.

\begin{table}[htb]

  \caption{Number of epochs compared with Root Mean Squared Error
    rounded to the nearest thousandth; all tests were run with EOS-USD
    as input.}

\label{tab:epochvsrmse}
\begin{tabular}{rl}
Epochs &   Root Mean Squared Error   \\
\hline
 5      & 0.523 USD              \\
 15     & 0.286 USD              \\
 25     & 0.260 USD              \\
 50     & 0.235 USD              \\
 100    & 0.229 USD              \\
\hline
\end{tabular}
\end{table}

\begin{figure*}[p]
\begin{minipage}{.05\textwidth}
\includegraphics[width=1.0\textwidth]{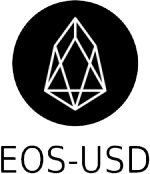}
\end{minipage}
\ \
\begin{minipage}{.4\textwidth}
\includegraphics[width=1.0\textwidth]{images/EOS-USD-prediction-model.pdf}
\end{minipage}
\ \
\begin{minipage}{.4\textwidth}
\includegraphics[width=1.0\textwidth]{images/EOS-USD-prediction-model-zoomed.pdf}
\end{minipage}

\begin{minipage}{.05\textwidth}
\includegraphics[width=1.0\textwidth]{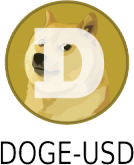}
\end{minipage}
\ \
\begin{minipage}{.4\textwidth}
\includegraphics[width=1.0\textwidth]{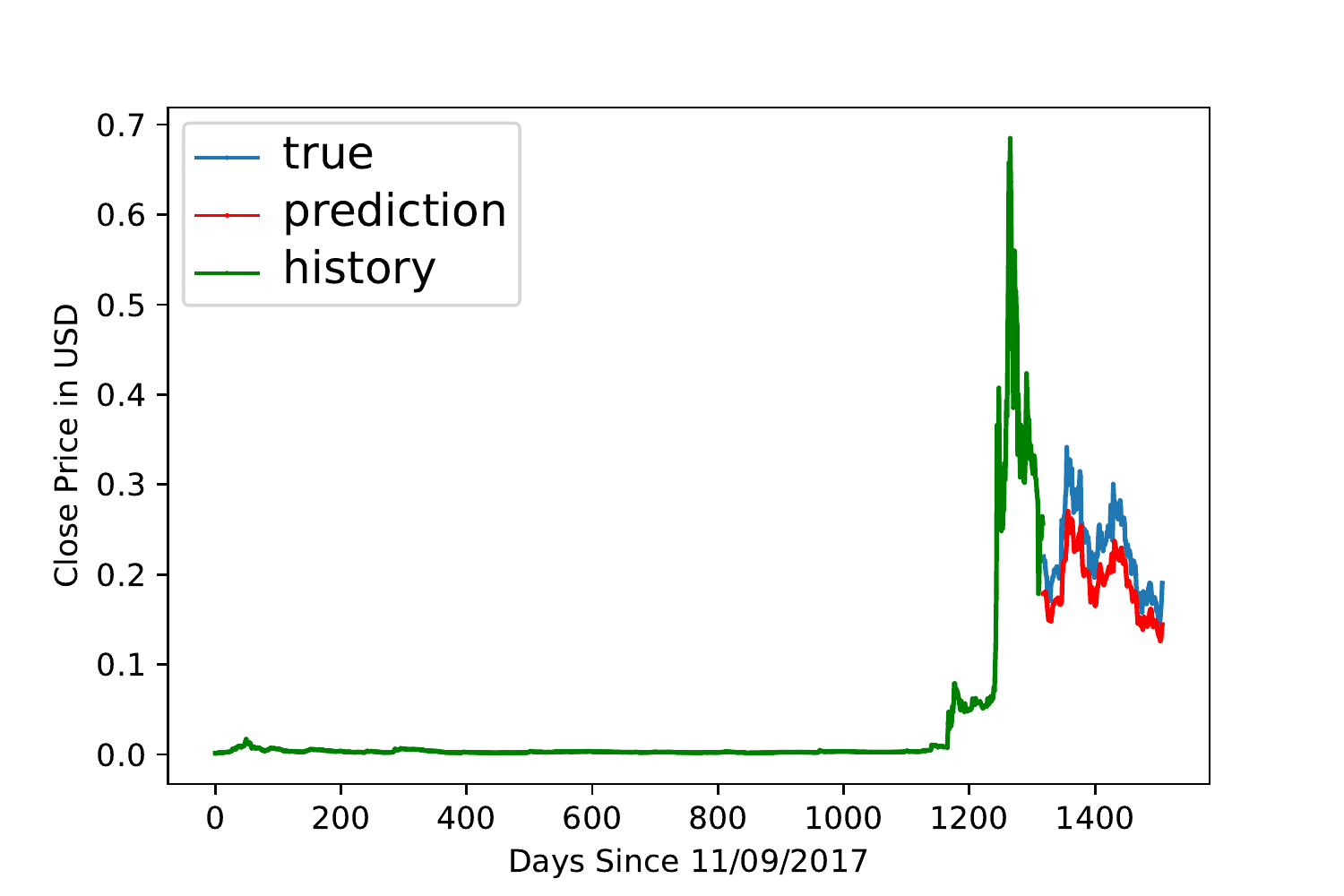}
\end{minipage}
\ \
\begin{minipage}{.4\textwidth}
\includegraphics[width=1.0\textwidth]{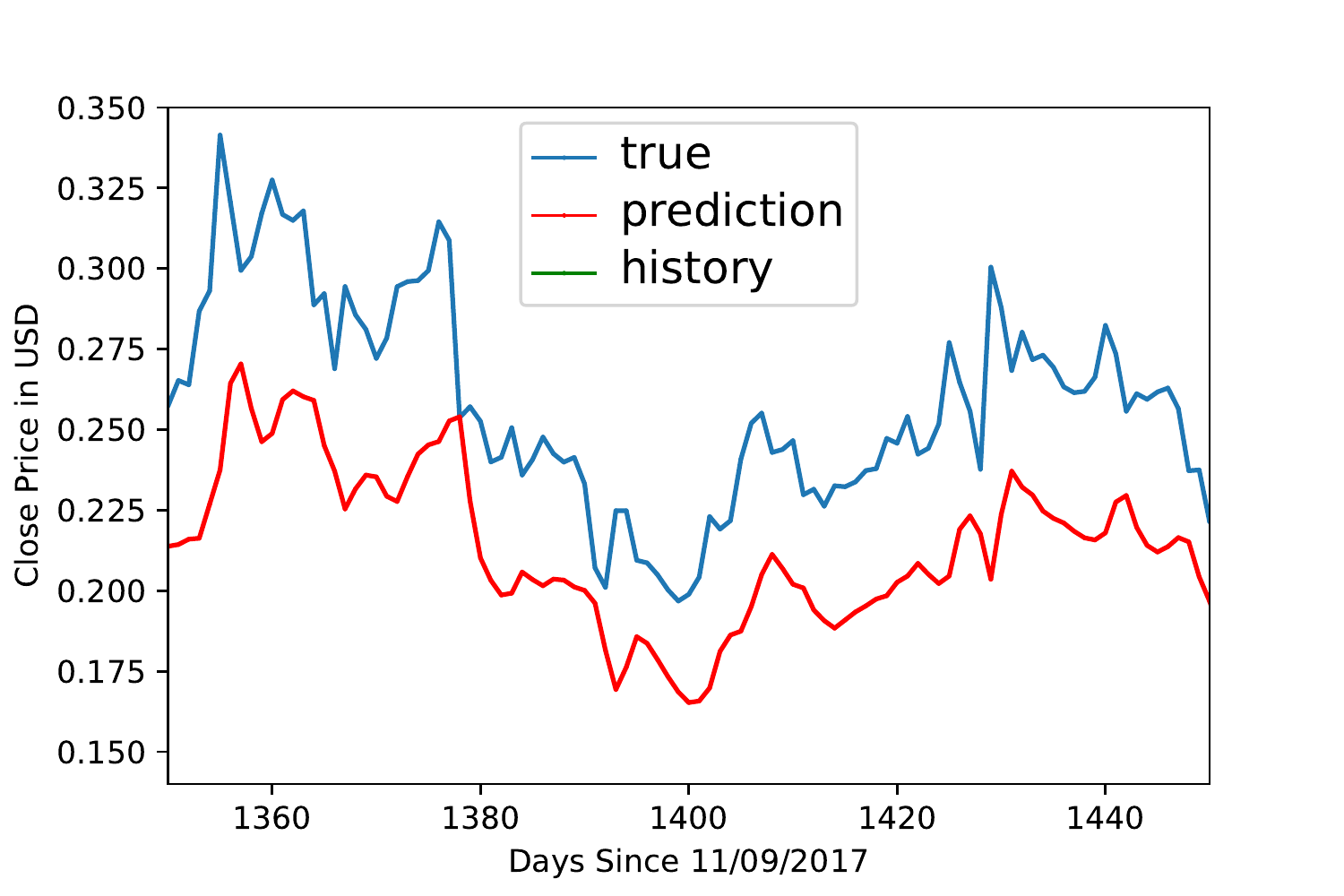}
\end{minipage}

\begin{minipage}{.05\textwidth}
\includegraphics[width=1.0\textwidth]{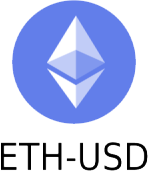}
\end{minipage}
\ \
\begin{minipage}{.4\textwidth}
\includegraphics[width=1.0\textwidth]{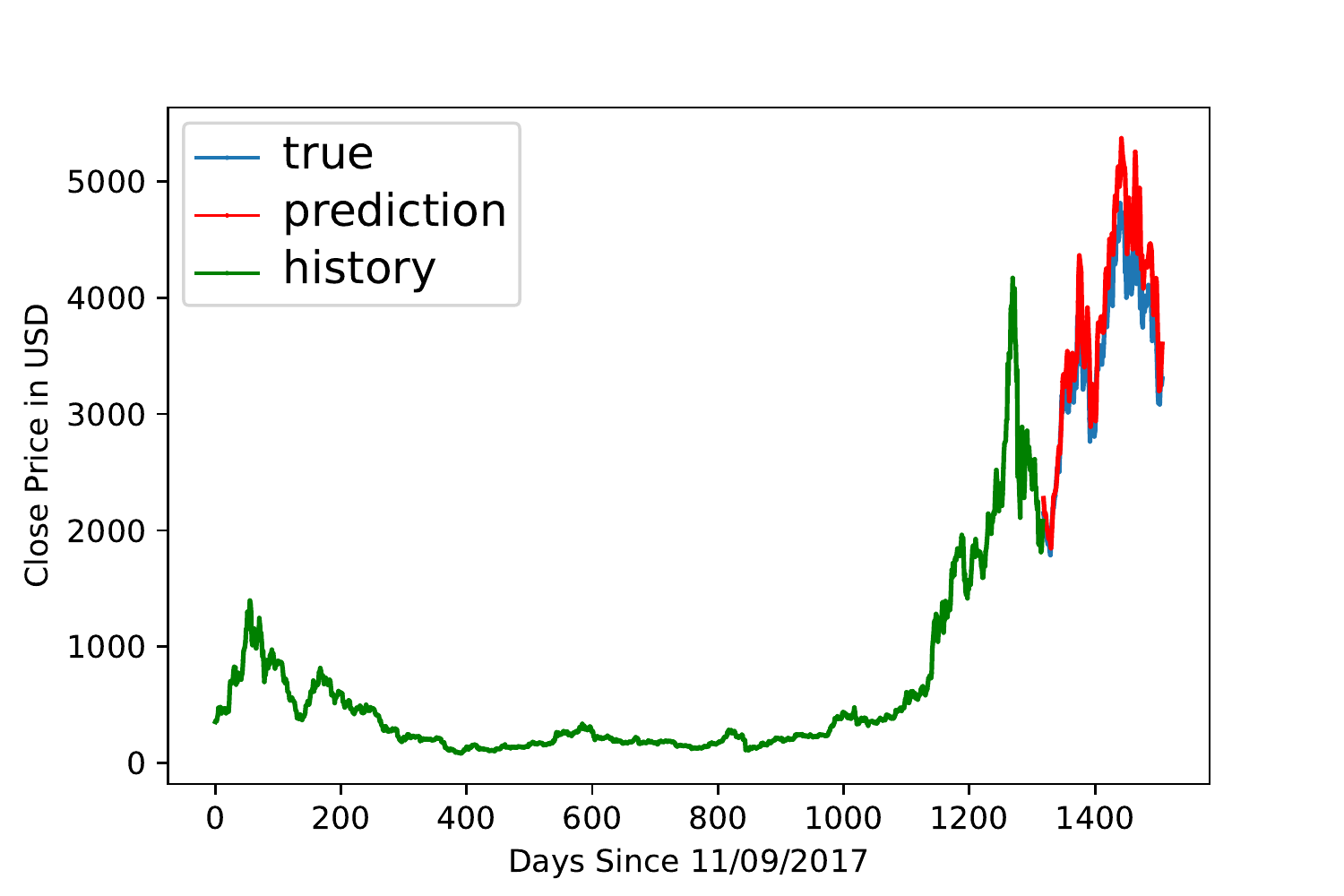}
\end{minipage}
\ \
\begin{minipage}{.4\textwidth}
\includegraphics[width=1.0\textwidth]{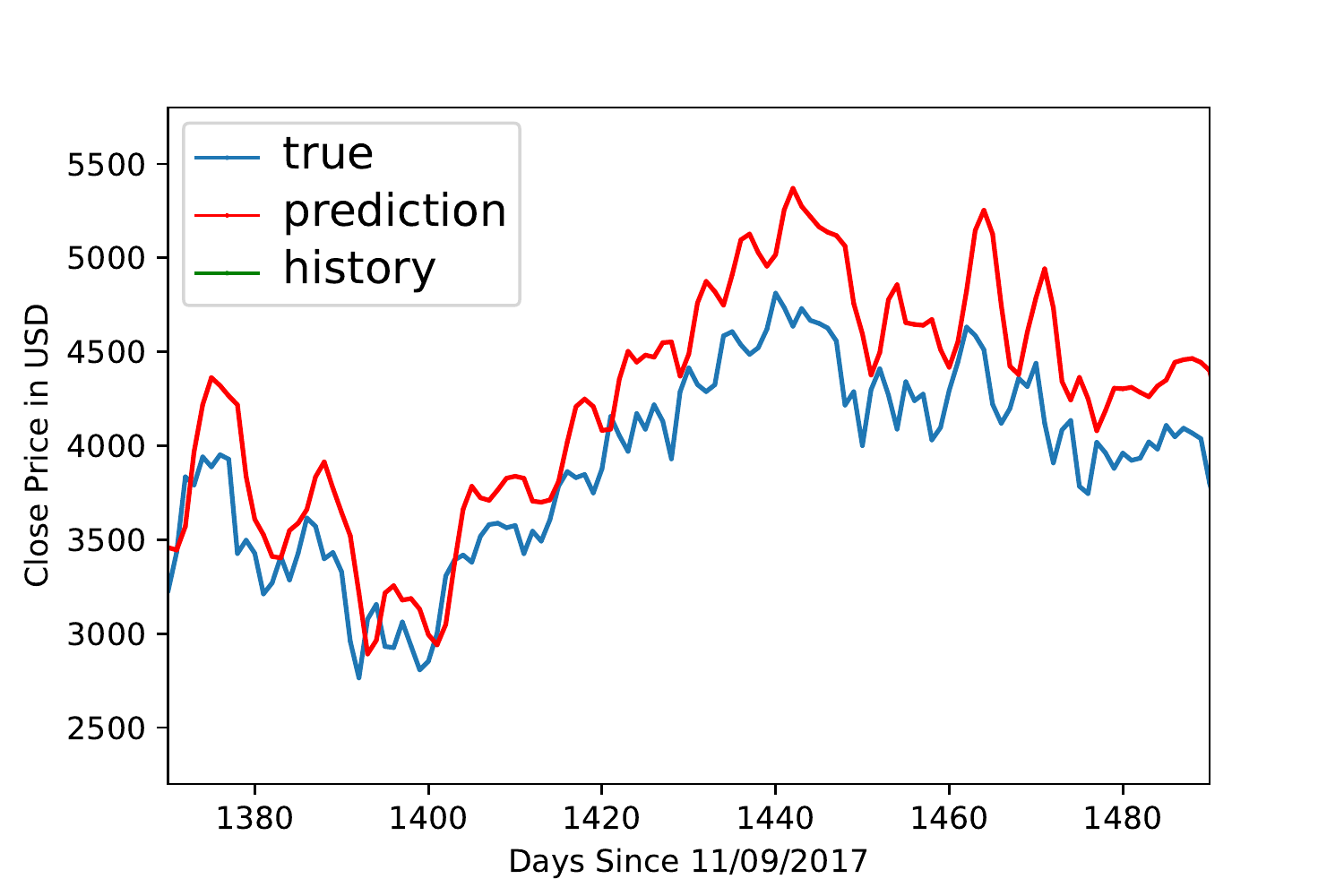}
\end{minipage}

\begin{minipage}{.05\textwidth}
\includegraphics[width=1.0\textwidth]{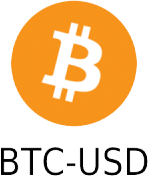}
\end{minipage}
\ \
\begin{minipage}{.4\textwidth}
\includegraphics[width=1.0\textwidth]{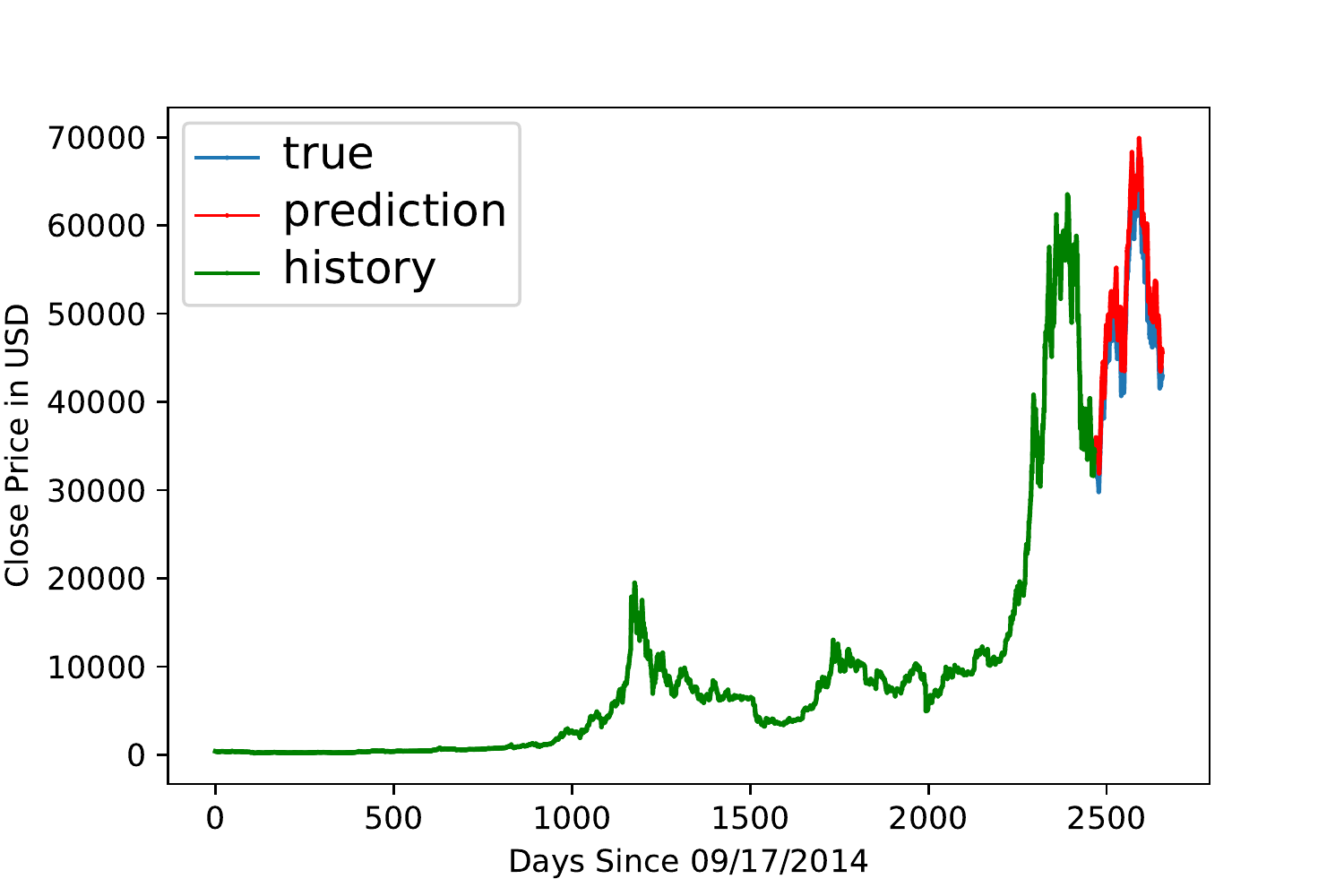}
\end{minipage}
\ \
\begin{minipage}{.4\textwidth}
\includegraphics[width=1.0\textwidth]{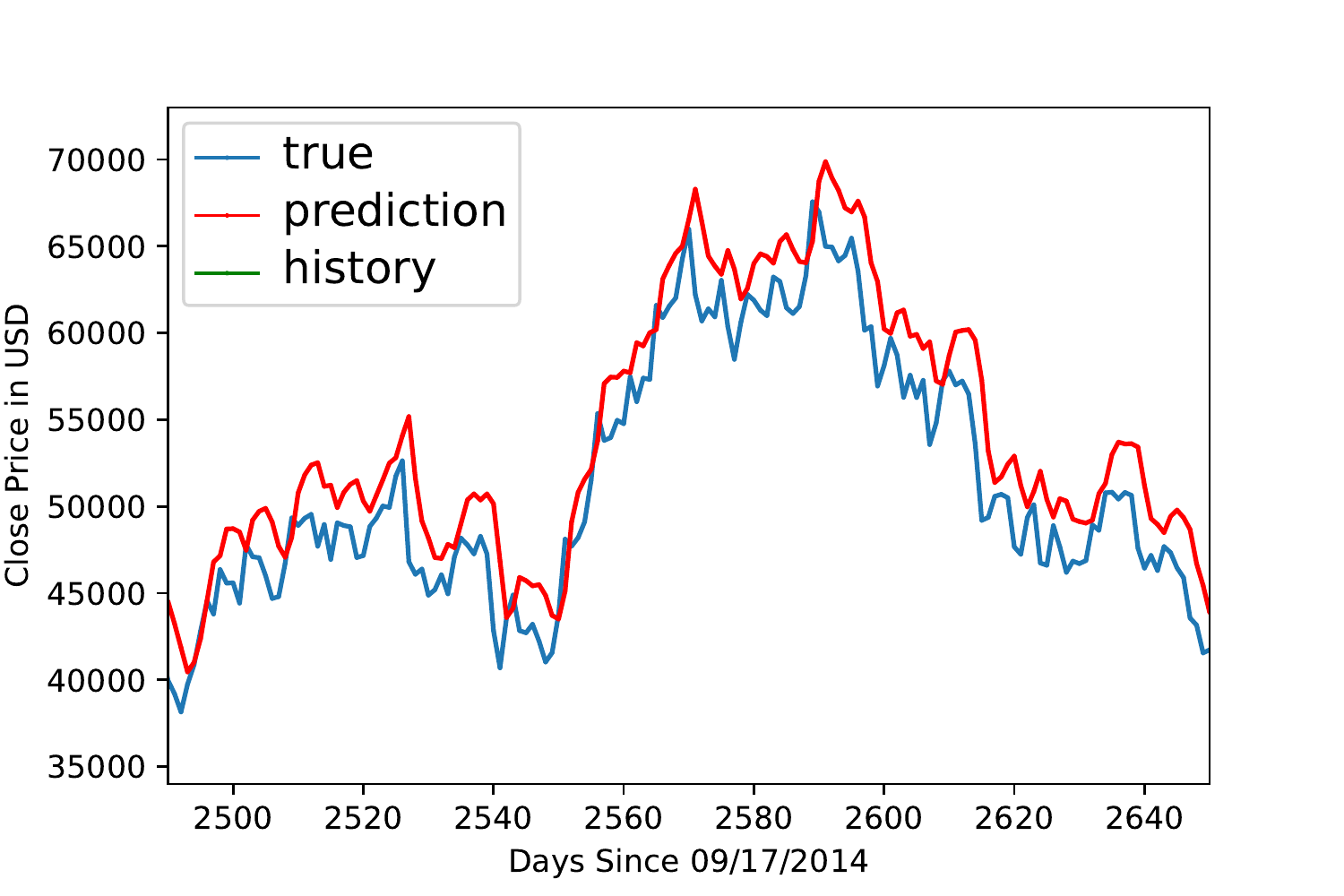}
\end{minipage}

\caption{EOS, Dogecoin, Ethereum, and Bitcoin prediction models.}
\label{fig:other}
\end{figure*}

Lastly, cryptocurrencies other than EOS such as Dogecoin, Ethereum,
and Bitcoin can be analyzed as well. Figure \ref{fig:other}
demonstrates the prediction models generated for these
cryptocurrencies. Dogecoin presents a model with predictions that are
more widely offset than the other coins, likely because most of the
training period encompasses a period of relative inactivity (no high
changes in price).

\section{Benchmark}

The benchmark is run within yfinance-lstm.ipynb located in
project/code \cite{c13}. The program ran on a 64-bit Windows 10 Home
Edition (21H1) computer with a Ryzen 5 3600 processor (3.6 GHz). It
also has dual-channel 16 GB RAM clocked at 3200 MHz and a GTX 1660
Ventus XS OC graphics card. Table \ref{tab:resource} lists these
specifications as well as the allocated computer memory during runtime
and module versions. Table \ref{tab:second} shows that the amount of
time it takes to train the 50 epochs for the LSTM is around 15
seconds, while the entire program execution takes around 16 seconds. A
StopWatch module was used from the package cloudmesh-common \cite{c10}
to precisely measure the training time.

In Table 3, the time column reports the length of the program phase in
seconds. Training time and prediction time do not perfectly add up to
overall time because the time it took to split data into train and
test sets is not part of the training or prediction
phases. Furthermore, the start times are similar because the entire
program's cells were run consecutively.
 
\begin{table}[htb]
  
  \caption{Benchmark details including the specifications and status
    of the computer at the time of program execution.}
  
\label{tab:resource}
\begin{tabular}{p{2cm}p{5cm}}
Attribute        & Value  \\
\hline
 cpu cores        & 6    \\
 cpu threads      & 12    \\
 cpu frequency    & 3600.0 MHz      \\
 mem.available    & 7.1 GiB \\
 mem.percent      & 55.3 \%  \\
 mem.total        & 16.0 GiB \\
 mem.used         & 8.8 GiB  \\
 python           & 3.9.5 (tags/v3.9.5:0a7dcbd, May  3 2021, 17:27:52) [MSC v.1928 64 bit (AMD64)]  \\
 python.pip       & 21.1.3   \\
 python.version   & 3.9.5 \\
 uname.processor  & AMD64 Family 23 Model 113 Stepping 0, AuthenticAMD   \\
 uname.system     & Windows \\
 uname.version    & 10.0.19043 \\
 \hline
 \end{tabular}
 \end{table}

\begin{table}[htb]
  
  \caption{Benchmark output which reports the execution time of
    overall program, training phase, and prediction phase.}
  
\label{tab:second}
\begin{tabular}{lrrll}
Name            &   Time   & Start               & OS Version \\
\hline
 Overall time    & 16.589 s & 2021-07-26 18:39:57 & Windows  10.0.19043, SP0 \\
Training time   & 15.186 s & 2021-07-26 18:39:58 & Windows 10.0.19043, SP0 \\
 Prediction time &  0.227 s & 2021-07-26 18:40:13 & Windows 10.0.19043, SP0 \\
 \hline
 \end{tabular}
 \end{table}
 
\section{Conclusion}

At first glance, the results look promising as the predictions have
minimal deviation from the true values (as seen in Figure
\ref{fig:zoomed}). However, upon closer look, the values lag by one
day, which is a sign that they are only viewing the previous day and
mimicking those values. Furthermore, the model cannot go several days
or years into the future because there is no data to run on, such as
opening price or volume. The experiment is further confounded by the
nature of stock prices: they follow random walk theory, which means
that the nature in which they move follows a random walk: the changes
in price do not necessarily happen as a result of previous
changes. Thus, this nature of stocks contradicts the very architecture
of this experiment because long short-term memory assumes that the
values have an effect on one another.

For future research, a program can scrape tweets from influencers'
Twitter pages so that a model can guess whether public discussion of a
cryptocurrency is favorable or unfavorable (and whether the price will
increase as a result).

\begin{acks}
  
Thank you to Florida A\&M University for graciously funding this
scientific excursion and the Miami Dade College School of Science for
this research opportunity. Work supported by Gregor von Laszewski was
supported by the NSF Grant \#1829704: CyberTraining: CIC:
CyberTraining for Students and Technologies from Generation Z.  Work
conducted by members of FAMU and Jacques Fleischer was supported, in
part, by NSF Grant called Florida Georgia Louis Stokes Alliance for
Minority Participation, with the subaward FAMU C-5083.

\end{acks}

\bibliographystyle{ACM-Reference-Format}
\bibliography{paper-blockchain}

\appendix

\section{Additional Material}

The following additional material is available:

\begin{description}

\item[Online Description] \cite{c14}

\item[Install documentation] \cite{c12}

\item[Python Notebook] yfinance-lstm.ipynb \cite{c13}

\item[Presentations] 
Presentations of this work were given at the
2021 FAMU-FGLSAMP Data Science and AI Research Experience for Undergraduates Presentation \cite{c15}
and as poster in the Miami Dade College School of Science 10th Annual STEM Research Symposium 2021 Poster \cite{c16}.

\end{description}

\end{document}